\documentclass[]{innovator}
\usepackage{enumitem}
\usepackage[toc,page,header]{appendix}
\usepackage[utf8]{inputenc}
\usepackage[T1]{fontenc}
\usepackage{hyperref}
\usepackage{url}
\usepackage{graphicx}
\usepackage{booktabs}
\usepackage{amsfonts}
\usepackage{nicefrac}
\usepackage{makecell}
\usepackage{microtype}
\usepackage{amsmath}
\usepackage{etoolbox}
\usepackage{lipsum}  
\usepackage{minitoc}
\usepackage{tablefootnote}
\usepackage{threeparttable}
\usepackage{wrapfig}
\usepackage{appendix}
\usepackage{multirow}
\usepackage{ulem}
\useunder{\uline}{\ul}{}
\usepackage{colortbl}
\usepackage{longtable}
\usepackage{color}
\usepackage{tikz}
\usetikzlibrary{arrows.meta,positioning,fit,calc}

\title{AutoAgent: Evolving Cognition and Elastic Memory Orchestration for Adaptive Agents}

\author[1,3]{Xiaoxing Wang$^{*}$}
\author[1,3]{Ning Liao$^{*}$}
\author[1,2]{Shikun Wei}
\author[1,3]{Chen Tang}
\author[1,3]{Feiyu Xiong}

\affiliation[1]{MemTensor (Shanghai) Technology Co., Ltd.}
\affiliation[2]{Shanghai Jiao Tong University}
\affiliation[3]{Institute for Advanced Algorithms Research, Shanghai.}

\abstract{
Autonomous agent frameworks still struggle to reconcile long-term experiential learning with real-time, context-sensitive decision-making. In practice, this gap appears as static cognition, rigid workflow dependence, and inefficient context usage, which jointly limit adaptability in open-ended and non-stationary environments. To address these limitations, we present AutoAgent, a self-evolving multi-agent framework built on three tightly coupled components: evolving cognition, on-the-fly contextual decision-making, and elastic memory orchestration.
At the core of AutoAgent, each agent maintains structured prompt-level cognition over tools, self-capabilities, peer expertise, and task knowledge. During execution, this cognition is combined with live task context to select actions from a unified space that includes tool calls, LLM-based generation, and inter-agent requests. To support efficient long-horizon reasoning, an Elastic Memory Orchestrator dynamically organizes interaction history by preserving raw records, compressing redundant trajectories, and constructing reusable episodic abstractions, thereby reducing token overhead while retaining decision-critical evidence.
These components are integrated through a closed-loop cognitive evolution process that aligns intended actions with observed outcomes to continuously update cognition and expand reusable skills, without external retraining. Empirical results across retrieval-augmented reasoning, tool-augmented agent benchmarks, and embodied task environments show that AutoAgent consistently improves task success, tool-use efficiency, and collaborative robustness over static and memory-augmented baselines. Overall, AutoAgent provides a unified and practical foundation for adaptive autonomous agents that must learn from experience while making reliable context-aware decisions in dynamic environments.
The codebase of AutoAgent will be released at \url{https://github.com/vicFigure/AutoAgent}.
}

\begin{document}
\maketitle
\begingroup
\renewcommand\thefootnote{\fnsymbol{footnote}}
\footnotetext[1]{These authors contributed equally to this work.}
\footnotetext[2]{Contact: 	Xiaoxing Wang (figure1\_wxx@sjtu.edu.cn)}
\endgroup


\clearpage
\tableofcontents
\clearpage
\clearpage

\section{Introduction}                                                                                                                                                                                                                                                                                                                                                                                                                                                                                                  
\begin{center}
\begin{minipage}{0.85\linewidth}
\small
\centering
``\textbf{All our cognition begins with experience.}''\par
\vspace{0.2em}
\raggedleft---Immanuel Kant
\end{minipage}
\end{center}


Agent systems that perceive a task, plan actions, interact with external tools or environments, and iteratively refine their behavior based on feedback have emerged as a practical paradigm for turning large language models (LLMs) into general-purpose problem solvers \cite{react2022,toolformer2023}. In contrast to single-turn question answering, agentic workflows decompose complex objectives into multi-step actions, such as information seeking, tool invocation (e.g., search, code execution), and interaction with software services. More recently, multi-agent frameworks further extend this paradigm by enabling role specialization and collaboration through communication and division of labor \cite{camel2023,autogen2023}. While demonstrating remarkable success across domains such as code generation, scientific discovery, and interactive task completion, many existing agent systems remain largely static in their core mechanisms. This static nature manifests in three intertwined limitations that constrain their autonomy, adaptability, and efficiency in open-ended environments.

\textbf{First}, agents operate with a fixed, human-specified cognition. Tools, peers, and contexts are described by static, hand-written functional prompts that cannot be updated through experience. This results in rigid and often sub-optimal decisions—agents may repeatedly misuse a tool because its documented preconditions are incomplete, or they may overlook a capable collaborator due to an outdated description of expertise. 
\textbf{Second}, their problem-solving logic is heavily reliant on human-designed, prior-specified workflows. Agents are often confined to fixed reasoning loops or rigid plans, which struggle to adapt when faced with novel situations or unexpected outcomes. This inflexibility is particularly problematic because agent decisions—on what to do next, or which tool or collaborator to invoke—are inherently non-stationary; the effectiveness of an action depends critically on the evolving context. 
\textbf{Third}, current systems suffer from inefficient context and memory management. While capable of processing long histories, they often treat past interactions merely as raw text appended to the prompt. This flat, linear growth leads to token redundancy, slowed reasoning, and difficulty in retrieving relevant information. Moreover, the lack of mechanisms to actively organize experience into structured knowledge (e.g., episodic memory or reusable skills) forfeits opportunities for long-term learning and efficient recall. 
These limitations—static cognition, workflow inflexibility, and context inefficiency—collectively constrain an agent's capacity for sustained autonomous learning and adaptation in dynamic, open-ended environments.

To address these limitations, we introduce a cohesive design philosophy centered on three integrated pillars: Evolving Cognition, On-the-fly Contextual Decision-Making, and Elastic Memory Orchestration. This framework, named AutoAgent, is designed to directly counter the aforementioned issues of static description, inflexible planning, and inefficient context management.

\textbf{First}, we formalize an agent's Evolving Cognition into two complementary aspects: Internal Cognition, comprising functional knowledge of tools and self-capabilities (skills), and External Cognition, modeling peer agents and anticipated environmental feedback. This structured representation moves beyond static prompts to form a dynamic, learnable knowledge base. 
\textbf{Second}, this cognition enables On-the-fly Contextual Decision-Making. By synthesizing its current cognition with the live context, an agent dynamically selects the most suitable action from a unified space, which includes Emic Actions (self-driven problem-solving using its internal capabilities) and Etic Actions (seeking external assistance or collaboration). This replaces rigid, pre-defined workflows with adaptive, moment-to-moment reasoning. 
\textbf{Third}, to make this process efficient and scalable, the Elastic Memory Orchestrator actively organizes the interaction context. It reduces token overhead, filters redundancies, and structures relevant history, thereby accelerating reasoning and ensuring that decisions are informed by concise, salient information rather than a bloated transcript.
\textbf{Finally}, these components are integrated through a Cognitive Evolution loop, where experiences from decision outcomes are analyzed to continuously update both internal and external cognition. Thus, the system operates as a self-reinforcing cycle: evolving cognition guides contextual decisions, decisions generate structured experiences via memory orchestration, and these experiences, in turn, refine the cognition—enabling sustained autonomous learning and problem-solving.

We evaluate AutoAgent across a range of benchmarks involving tool use, multi-agent collaboration, and long-horizon reasoning. Results demonstrate consistent and significant improvements in task completion rate, tool-use efficiency, and collaborative robustness compared to static agents and prior memory-augmented systems.
In summary, this work makes the following key contributions:

\textbf{1) Evolving Cognition as a Learnable Agent State.} We introduce a structured, dual-faceted cognition—comprising internal knowledge of tools and skills, and external cognition of peers and environment—that serves as an explicit, updatable agent state. Moving beyond static prompts, this representation is continuously refined through interaction, directly enabling more reliable tool selection and informed collaboration.

\textbf{2) A Unified Framework for On-the-fly Contextual Decision-Making.} We formulate agent operation as an iterative loop of cognition- and context-driven action selection, unifying emic (self-driven) and etic (help-seeking) actions within a single space. This design replaces pre-defined workflows with adaptive, moment-to-moment reasoning, offering a consistent interface for execution, logging, and learning.

\textbf{3) Elastic Memory Orchestration for Efficient Long-horizon Reasoning.} To tackle context inefficiency, we propose an Elastic Memory Orchestrator that dynamically compresses history, summarizes episodes, and distills reusable skills. This active management reduces token overhead, accelerates reasoning, and ensures decisions are informed by concise, salient experience rather than raw, growing transcripts.

\textbf{4) Closed-loop Cognitive Evolution from Practice.} We develop a self-evolution mechanism that closes the loop between action outcomes and agent cognition. By analyzing trajectories and aligning intent with results, the system continuously updates its cognitive models and skill library without external retraining, allowing the agent to progressively adapt in non-stationary environments.

Together, these contributions present AutoAgent: a unified and self-improving framework that bridges evolving cognition, contextual decision-making, and elastic memory management for autonomous problem-solving.

\section{Related Work}
We review prior work along four dimensions closely related to AutoAgent: tool-using language agents, multi-agent collaboration, memory-augmented agents, and self-improving or reflective agents.

\textbf{Tool-using Agents.} 
A growing line of work augments LLMs with external tools (e.g., search, code execution, APIs) to improve factuality and task solvability.
Representative systems combine reasoning with tool invocation via explicit action traces (e.g., ReAct)~\cite{react2022}, teach models to call tools through self-supervision (e.g., Toolformer)~\cite{toolformer2023}, or route queries to specialized modules/tools (e.g., MRKL)~\cite{mrkl2023}.
Recent advances focus on \textit{when} to invoke tools: MeCo introduces meta-cognition as a proxy for LLMs' self-assessment of internal capability, using representation-space probes to dynamically decide tool necessity and avoid unnecessary calls~\cite{li-etal-2025-adaptive}.
Other systems focus on tool-use training datasets, tool-use evaluation and large tool libraries (e.g., ToolBench~\cite{toolbench2023})~\cite{toolbench2023,chen-etal-2024-towards-tool,you-etal-2024-mumath,ye-etal-2025-toolhop,theuma-shareghi-2024-equipping}.
While effective, these approaches—including meta-cognitive triggering—treat tool properties (reliability, preconditions, costs) as static and do not maintain an explicit, updatable belief model that evolves through interaction experience.

\textbf{Multi-agent Collaboration.} 
Multi-agent frameworks leverage communication and role specialization to tackle complex tasks.
CAMEL explores structured dialogs between agents~\cite{camel2023}, while AutoGen provides a programmable multi-agent conversation framework~\cite{autogen2023}.
AnyMAC~\cite{wang-etal-2025-anymac} reformulates multi-agent coordination as sequential next-agent prediction rather than fixed graph topologies, enabling dynamic agent reuse and flexible context selection from any prior step to support adaptive communication pipelines.
The Explain-Analyze-Generate (EAG) framework~\cite{gu-etal-2025-explain} decomposes complex reasoning into a sequential pipeline of specialized subtasks (explanation, analysis, generation) to mitigate error propagation inherent in parallel debate paradigms.
Most existing systems focus on communication protocols, orchestration, and task decomposition strategies, but leave open the question of how agents should continuously update their cognition about peers (capabilities, expertise, and reliability) based on interaction evidence.

\textbf{Memory-augmented Agents.} 
To cope with long-horizon tasks, several systems introduce explicit memory modules.
MemGPT frames memory as a managed resource with mechanisms for moving information between short-term context and long-term storage~\cite{memgpt2023}.
Generative Agents demonstrate how event memories and reflections can support sustained behavior in simulated environments~\cite{generativeagents2023}.
Recent work explores autonomous memory augmentation: MemInsight~\cite{salama-etal-2025-meminsight} enables agents to proactively identify critical information and generate semantic attributes for memory entries, improving retrieval accuracy through structured representation.
However, memory is often treated as passive storage or limited to single-interaction augmentation; less attention is paid to distilling \textit{repeated experiences across interactions} into reusable skills and integrating them into an agent's cognition and action selection.

\textbf{Self-evolving Agents.} 
Reflexion shows that language agents can improve by reflecting on failures and incorporating verbal feedback into subsequent trials~\cite{reflexion2023}.
EvolveSearch~\cite{zhang-etal-2025-evolvesearch} introduces an iterative self-evolution framework that alternates between RL exploration and SFT optimization, bootstrapping agent capabilities by leveraging high-reward rollouts as self-generated supervision without human-annotated data.
Complementary directions include explicit search over reasoning trajectories (e.g., Tree of Thoughts)~\cite{tot2023} and embodied lifelong learning agents that acquire skills through interaction (e.g., Voyager)~\cite{voyager2023}.
These methods—including iterative self-evolution—remain largely episodic and task-specific, and do not explicitly couple reflection with a persistent cognitive state, elastic memory compression, and multi-agent collaboration.

\begin{figure}[t]
    \centering
    \includegraphics[width=0.9\textwidth]{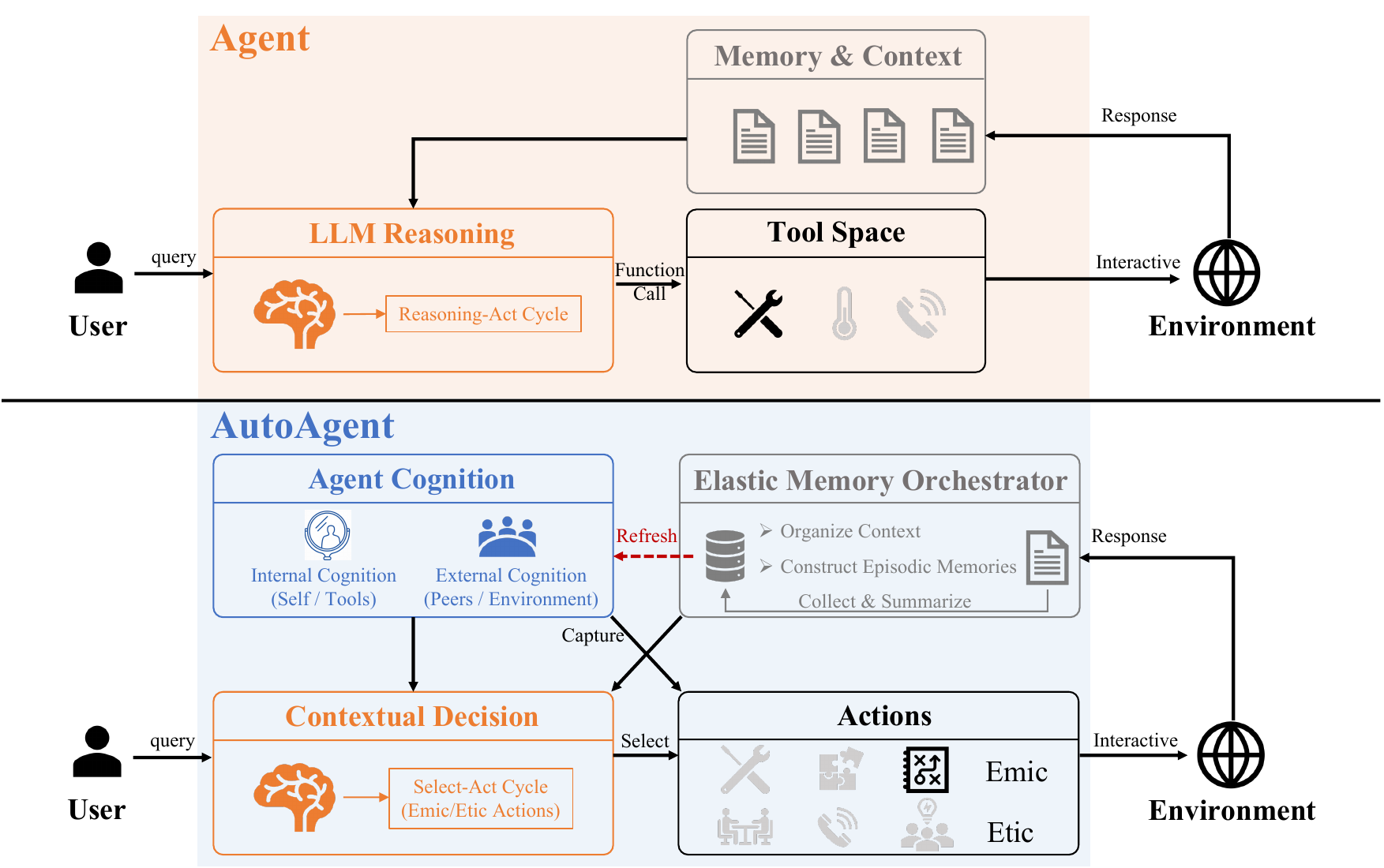}
    \caption{The AutoAgent architecture. The \textbf{Execution Cycle} (solid black arrows) handles real-time task progression. The \textbf{Evolution Cycle} (dashed red arrows) drives long-term adaptation. The Elastic Memory Orchestrator couples both cycles by managing experience.}
    \label{fig:overview}
\end{figure}

\section{The AutoAgent Framework: A Unified Architecture for Self-Evolution}

This chapter presents the integrated architecture of AutoAgent, which materializes the self-evolution paradigm introduced in Section 1. Moving beyond a collection of isolated mechanisms, AutoAgent is designed as a cohesive system where components interact within a closed-loop process to enable continuous adaptation. We first articulate the overarching design philosophy (4.1), then detail the static architectural blueprint (4.2), and finally explain the dynamic interactions that realize self-evolution (4.3).

\subsection{Design Philosophy: The Self-Evolution Loop}
Contemporary agent frameworks often treat an agent's knowledge base—its understanding of tools, peers, and task contexts—as a static, human-specified artifact. This results in the brittle and inefficient behaviors outlined in the introduction. In contrast, AutoAgent is founded on the principle that an agent's operational knowledge must be continuously verifiable and updatable through its own practice. We formalize this principle as the Self-Evolution Loop, a recursive paradigm structured around four core functions: Cognition that maintain structured, descriptive knowledge; Decision that select contextually appropriate actions using current cognition; Memory that organize the history of actions and outcomes; Evolution that analyze organized experience to refine cognition and memory.
The key insight is that these functions form a closed cycle: action (Decision) generates experience (Memory); analysis of experience (Evolution) produces refined knowledge (Cognition); and updated knowledge enables more effective future actions. This loop intrinsically unifies “acting” with “learning,” providing the foundational blueprint for the AutoAgent architecture.

\subsection{Architectural Blueprint: Components and Their Roles}
The Self-Evolution Loop is instantiated in AutoAgent through four principal components, whose relationships are depicted in Figure~\ref{fig:overview}. Each component assumes a distinct, critical role in realizing the system's adaptive capability.

\noindent\textbf{1. Cognition Layer: The Evolvable Knowledge Base.} This component maintains the agent's structured understanding as two interrelated facets: Internal Cognition (functional descriptions of tools and self-capabilities) and External Cognition (descriptive profiles of peer agents and environmental dynamics). It serves as the system's long-term, updatable repository of know-how.

\noindent\textbf{2. Contextual Decision Engine: The Real-Time Executor.} This engine operationalizes the agent's moment-to-moment problem-solving. It engages in an atomic Select-Execute cycle, where it chooses an action—either an Emic Action (self-reliant) or an Etic Action (collaborative)—by reasoning over the current context (provided by memory) and the actionable knowledge from the Cognition Layer.

\noindent\textbf{3. Elastic Memory Orchestrator: The Experience Manager.} This component is responsible for the efficient organization of the agent's interaction history. It dynamically compresses raw experience traces, summarizes coherent event sequences into episodic memories, and retrieves task-relevant information to construct concise working contexts. It is the system's central hub for experience.

\noindent\textbf{4. Cognitive Evolution Module: The Self-Improvement Engine.} This module performs meta-cognitive analysis. By examining trajectories of experiences curated by the Memory Orchestrator, it identifies discrepancies between agent intentions and actual outcomes. It subsequently formulates precise updates—expressed as revisions to descriptive text—to correct and enrich the knowledge stored in the Cognition Layer and to optimize the Memory Orchestrator's strategies.

\subsection{System Dynamics: Realizing Self-Evolution through Interaction}
The adaptive behavior of AutoAgent emerges from the interaction between its components, which operate through two tightly coupled cycles: a fast Execution Cycle and a slower Evolution Cycle (see Figure~\ref{fig:overview}).

\textbf{The Execution Cycle: Enabling Contextual Adaptation.} 
This cycle handles real-time task progression and directly addresses the problem of inflexible, pre-defined workflows.
1) The Contextual Decision Engine prepares for a decision by querying the Cognition Layer for relevant tool/peer descriptions and receiving from the Elastic Memory Orchestrator a dynamically assembled, compressed history relevant to the current step.
2) Using this integrated information, the Engine performs its Select operation, making an immediate, context-sensitive action choice. This on-the-fly decision-making allows the agent to dynamically replan, bypassing rigid workflows.
3) After Execute yields an outcome, the full record (intention, action, parameters, result) is sent as a raw experience trace back to the Memory Orchestrator for ingestion.
In this cycle, the Memory Orchestrator acts as an efficient context provider, actively mitigating token overhead and information redundancy to solve the challenge of inefficient context management during reasoning.

\textbf{The Evolution Cycle: Driving Continuous Learning.} 
This cycle facilitates long-term competence growth and directly attacks the root cause of static, biased cognitive descriptions.
1) The Elastic Memory Orchestrator periodically supplies structured experience data—such as summarized trajectories or episodic memories—to the Cognitive Evolution Module.
2) The Evolution Module conducts retrospective analysis, comparing recorded action intentions with observed outcomes to diagnose failures, successes, and patterns.
3) Based on this analysis, it generates explicit updates. These may include correcting a tool's precondition in the Internal Cognition, refining a peer's expertise description in the External Cognition, or proposing a new skill template derived from a successful action sequence.
4) These updates are committed back to the Cognition Layer and may also refine the compression/retrieval policies of the Memory Orchestrator itself.
Here, the Memory Orchestrator serves as a structured experience curator, transforming raw interaction data into a form suitable for high-level meta-cognitive analysis.

The two cycles are not isolated; they are synergistically integrated through the shared Elastic Memory Orchestrator. The Execution Cycle produces the raw experiential data that fuels the Evolution Cycle. In turn, the Evolution Cycle refines the cognitive knowledge that guides future executions. This creates a virtuous, self-reinforcing loop: better cognition leads to more effective decisions, which generate higher-quality experience for learning, which produces even better cognition.

This architectural integration is the cornerstone of AutoAgent's self-evolution capability. It provides a systematic pathway for an agent to start with initial, potentially imperfect descriptions and progressively ground its knowledge in operational reality, thereby achieving sustained improvement in autonomy, efficiency, and robustness. The following chapters delve into the detailed design of each core component. 

\section{Agent Cognition}


We introduce cognition as a first-class, structured representation within an agent system. It serves as the essential interface between the reasoning capacity of a Large Language Model (LLM) and the agent's external action space. In AutoAgent, any description that mediates the agent's interaction with the world—including tool specifications, skill definitions, and models of peer agents—constitutes its cognition. This perspective reframes cognition from a static collection of prompt fragments into an evolving, practice-driven interface that enables an agent to understand its capabilities, comprehend its environment, and ultimately refine its ability to act effectively.

\textbf{The Limitation of Static Descriptions.} 
Prevailing agent frameworks predominantly rely on human-defined, static descriptions. Tools are introduced via fixed schemas or docstrings, and other agents are represented by predefined role prompts. These descriptions are inherently incomplete and prone to bias: they may omit critical preconditions, inaccurately represent reliability, or fail to specify contextual appropriateness. Consequently, agents operating with such static cognition often make suboptimal decisions—repeatedly selecting ineffective tools, seeking help from ill-suited peers, or struggling to adapt as the task or tool ecosystem evolves.

\begin{figure}[t]
\centering
\vspace{-15pt}
\includegraphics[width=0.99\textwidth]{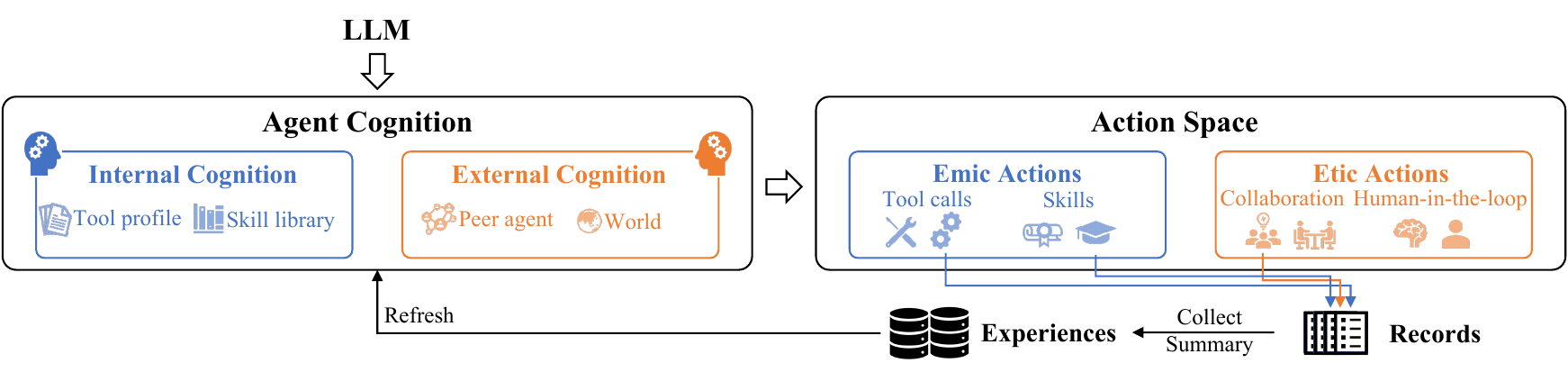}
    \caption{Cognition as the interface between the LLM and the action space. Tool calls, collaboration, and self-driven generations are treated as actions; their descriptions and beliefs are part of cognition and are updated from practice through execution outcomes and memory updates.}
    \label{fig:cognition-loop}
\end{figure}
\textbf{Cognition for Dynamic Action Selection.} 
We posit that an agent's core problem is the ongoing selection of the right action from a heterogeneous space encompassing tool calls, LLM-based generation (emic actions), and collaboration requests (etic actions). Cognition provides the necessary substrate for this selection. It formally represents the meaning, preconditions, and empirical knowledge associated with each action type. For instance, it encodes not just a tool's API but also learned insights about its successful application; it models a peer not just by a static role but by an evolving profile of observed expertise. Thus, cognition actively mediates how the LLM interprets the current context to produce an executable action, and it is continuously updated from interaction outcomes to improve future decisions. This closed-loop, practice-driven evolution is fundamental to moving beyond the limitations of biased, human-authored prior specifications.

To systematically capture an agent's knowledge, we structure its cognition into two complementary facets: Internal Cognition, which pertains to the self, and External Cognition, which models the world beyond the self. This structured duality provides clear pathways for learning and adaptation.

\subsection{Internal Cognition: Knowledge of Self}
Internal Cognition constitutes an agent's structured self-model, serving as a formal representation of its inherent capabilities and accessible resources. It addresses the fundamental question of what an agent itself can do, thereby forming the foundation for initiating emic actions—those instances where the agent relies on its own faculties to solve problems. 

This cognitive facet primarily encompasses two intertwined elements: tool profiles and a skill library. A tool profile extends beyond a static API signature to include an evolving, practical understanding of each tool’s purpose, validated preconditions for successful use, common failure patterns, and output characteristics. For instance, an agent's profile for a web\_search tool may be refined over time to reflect its particular efficacy for factual queries as opposed to subjective comparisons. Concurrently, the skill library houses reusable procedural templates—generalized solutions distilled from recurrent successful experiences, such as “formatting a bibliographic entry” or “parsing a specific error log.” Each skill is stored with its associated intent, triggering conditions, and an executable reference. 

Collectively, Internal Cognition functions as a dynamic, queryable self-knowledge base. During decision-making, it enables the agent to efficiently assess its own capacity, select the most appropriate internal resource, and apply it with informed parameterization, thereby enhancing autonomy, precision, and operational efficiency.

\subsection{External Cognition: Knowledge of the World}
External Cognition forms an agent's model of the environment external to itself, capturing knowledge about other actors and the general dynamics of task execution. It seeks to answer contextual questions regarding who else is available, what they are capable of, and how the world tends to respond to various actions. This representation is essential for enabling strategic etic actions, such as targeted collaboration. 

External Cognition is principally composed of dynamic peer agent models and contextual feedback estimates. A peer agent model evolves through interaction, building a profile of another agent’s specialized capabilities, observed reliability, and typical response patterns. This allows collaboration to shift from a broadcasted request to a directed consultation with the most qualified expert. Simultaneously, the agent develops estimates of environmental feedback, building an understanding of common action-outcome relationships within its operational domain to better anticipate results and plan more robustly. The overarching role of External Cognition is to ground the agent within its social and operational context. By maintaining an updated model of peers and environmental dynamics, it facilitates sophisticated social reasoning, intelligent task delegation, and the effective leveraging of collective expertise within a multi-agent ecosystem.

\section{Contextual Decision-Making}

Conventional agent approaches often rely on a priori planning, decomposing a task into a linear sequence of actions before execution begins. However, in dynamic environments where tool behaviors are stochastic and task states evolve unpredictably, such predetermined workflows can rapidly become suboptimal or obsolete. A static plan cannot adequately accommodate unexpected execution feedback, newly emergent constraints, or unforeseen opportunities for collaboration that arise during the reasoning process. This inherent brittleness highlights the necessity for a more adaptive paradigm. We posit that true agent autonomy is realized not through rigid adherence to a pre-specified script, but through contextual decision-making: an iterative process of making dynamic, moment-to-moment choices that are continuously conditioned on the agent's evolving cognition and the immediate task context. This shift from static planning to contextual adaptation empowers agents to navigate uncertainty, recover from failures, and opportunistically integrate new information, forming the foundation for robust and flexible problem-solving.

\subsection{The Atomic Decision Cycle: Select, Execute, Update}
Inspired by the reasoning-and-acting paradigm~\cite{react2022}, AutoAgent formalizes agent behavior as a repeating atomic decision cycle composed of three phases: \textbf{Select}, \textbf{Execute}, and \textbf{Update}. This cycle operationalizes contextual decision-making into a concrete, iterative process.

The \textbf{Select} phase constitutes the core of the decision-making process. Leveraging the current working context (curated by the Elastic Memory Orchestrator) and its own cognition, the agent's LLM is tasked with choosing the single most suitable next action. We deliberately isolate Select as an explicit, inspectable operation. This separation clarifies the boundary between general task reasoning and the specific reasoning required for action choice, preventing the entanglement often found in monolithic "thought-action" generation patterns. Making selection explicit yields several key benefits: it establishes a clear control point, enables consistent logging and supervision of decisions, and provides a well-defined interface for subsequent cognitive updates based on action outcomes.

The \textbf{Execute} phase, following selection, carries out the chosen action within the environment. Execution may involve invoking an external tool, performing a self-driven generation, or sending a collaboration request to another agent. This phase yields a concrete, observable outcome—such as a successful result, an error, or a peer agent's response.

The \textbf{Update} phase then integrates this outcome into the agent's state. The immediate result is appended to the task context, enriching the historical record for future reasoning. More profoundly, the outcome serves as critical feedback for long-term adaptation. It fuels the self-evolution mechanism, prompting potential refinements to the agent's cognition (e.g., updating tool reliability estimates or peer expertise models) and influencing the structuring of episodic memory and skill distillation within the Elastic Memory Orchestrator. This closed-loop design ensures a tight coupling between action and learning, where every execution informs and improves subsequent decisions.

\subsection{A Unified Action Space: Emic and Etic Actions}

To operationalize the Select phase, we define a unified action space that categorizes all possible agent behaviors along a key dimension: whether the action relies on its internalized capabilities or deliberately seeks to leverage external resources. This leads to a bipartite taxonomy of \emph{Emic Actions} and \emph{Etic Actions}.

\textbf{Emic Actions} embody an agent's capacity for autonomous problem-solving. These are actions performed by relying solely on the agent's internal resources, primarily its LLM's generative power and the knowledge encapsulated within its Internal Cognition—namely, its understanding of tools and repertoire of learned skills. This category unifies what might otherwise be treated separately. It includes: (1) self-driven generation, such as producing a chain of thought, drafting code, or summarizing information; and (2) the invocation of a known tool or application of a learned skill, as these are executed based on the agent's internalized model of its own capabilities. The defining characteristic of an Emic action is self-reliance; the agent operates under the assumption that the necessary competency resides within its cognitive model.

\textbf{Etic Actions}, in contrast, represent an agent's capacity for strategic help-seeking and collaboration. These are actions taken to solicit information, effort, or perspective from an entity external to the agent's direct control—most commonly another intelligent agent within the system. An Etic action is fundamentally a social or interactive move, such as querying a specialized peer for information, delegating a sub-task, or requesting verification. The selection and parameterization of an Etic action are directly guided by the agent's External Cognition—its maintained descriptive knowledge of peer capabilities, expertise, and reliability, which is refined through interaction. This framework elevates inter-agent interaction from an ad-hoc protocol to a first-class action type within the core decision cycle.

This dichotomous design of the action space serves multiple purposes. First, it aligns naturally with the structure of the agent's dual-faceted cognition (Internal vs. External), making the decision logic more transparent and tractable. Second, it provides a coherent foundation for the self-evolution mechanism to learn effective decision policies, such as when to rely on internal capabilities versus when to seek external assistance. Finally, it unifies traditionally distinct paradigms—single-agent tool use and multi-agent collaboration—under a singular decision-making framework, simplifying the agent architecture and enabling more general learning from heterogeneous experiences.

\section{Elastic Memory Orchestration}

\begin{wrapfigure}{r}{0.58\textwidth}
    \centering
    \vspace{-0.8em}
    \includegraphics[width=0.9\linewidth]{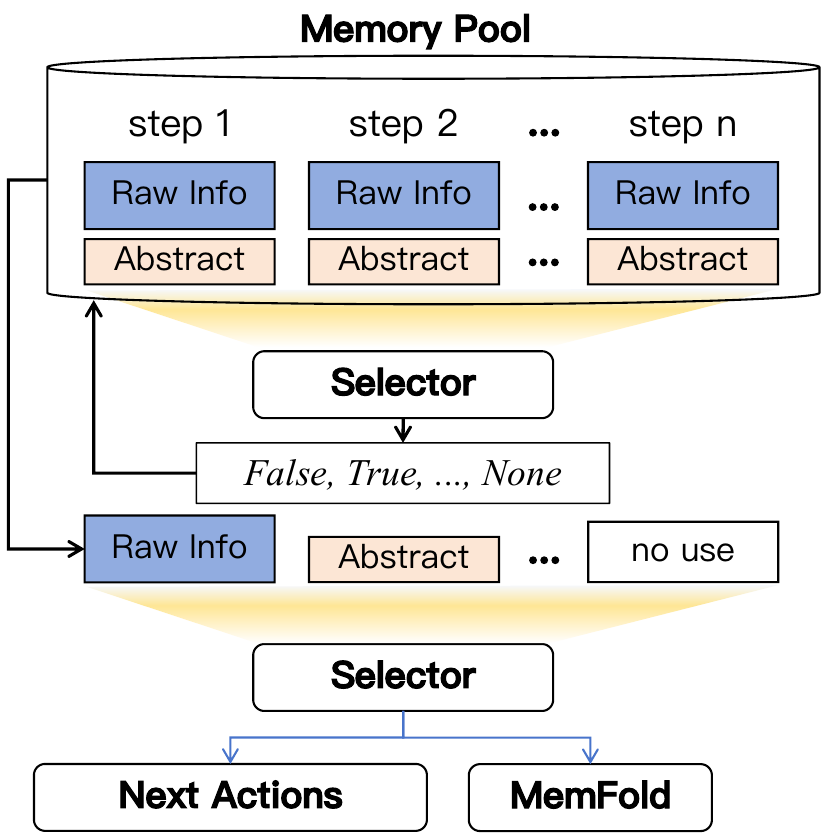}
    \caption{The framework of the proposed Elastic Memory Orchestration (EMO) module. At each step of history, action information will be preserved in the Memory Pool in two forms: the complete and lossless raw data, and the summarized action abstracts. At the beginning of each current step, the EMO module first concatenates the action abstracts from all historical steps as input to the selector. The selector determines whether the actions in each round should use the raw information (False), the summarized information (True), or discard the corresponding step’s information (None). After this round of judgment, the historical action information is reorganized accordingly. Subsequently, the selector further decides whether to fully summarize multiple historical steps by executing MemFold, or proceed with other actions such as web search, and so on.}
    \label{fig:EMO}
    \vspace{-0.8em}
\end{wrapfigure}
A core challenge in sustaining long-horizon agent autonomy is the efficient management of an ever-growing interaction history. Naive approaches that concatenate raw observations, actions, and outcomes into a linear prompt are unsustainable, inevitably exceeding the context window limits of the underlying language model and introducing significant computational overhead. While some existing methods address this by applying lossy compression to past steps, they often rely on a single, irreversible transformation. This creates a fundamental tension: excessive compression sacrifices details crucial for precise future reasoning, while retaining full fidelity bloats the context and slows cognition. Moreover, the relevance of historical information is not static; details critical for one decision may be superfluous for another. 
Therefore, an agent requires not merely compression, but intelligent, elastic memory organization—a system that dynamically adapts the representation of the past to meet the specific needs of the present. The Elastic Memory Orchestrator (EMO) is designed to fulfill this role, as depicted in Fig.~\ref{fig:EMO}. Its primary objectives are twofold: (1) to drastically reduce token consumption and accelerate reasoning by actively filtering and summarizing redundant or tangential information, and (2) to enhance the quality of contextual decision-making by ensuring that the working memory presented to the agent is both concise and rich in task-relevant details. This enables sustained, efficient, and context-aware operation over extended task sequences.

\subsection{Step-wise Action Memory: Selective Compression and Dynamic Retrieval}

To achieve elasticity, our memory orchestrator operates on a stepwise granularity, treating each iteration of the Select-Execute-Update cycle as a distinct memory unit. For each completed action, the EMO performs a two-stage process: compression and dynamic assembly.

\textbf{Compression Stage:} Post-execution, the detailed record of a step—comprising the agent's selection rationale, the specific action and its parameters, and the execution outcome—is processed to generate a compact summary. This summary retains the step's semantic essence and outcome while discarding verbose reasoning traces or redundant details. Consequently, for every historical step $i$, the memory system maintains a dual representation: the raw, complete record $R_i$ and its corresponding compressed summary $C_i$.

\textbf{Dynamic Assembly Stage:} Before a new step begins, the EMO dynamically constructs the working context for the agent. It presents the selector (a learned or heuristic module) with the sequence of compressed summaries $[C_1, ..., C_{t-1}]$ from previous steps. The selector's task is to judge, for each step $i$, which representation is most appropriate for the upcoming decision: whether the full detail of $R_i$ is required, whether the summary $C_i$ suffices, or whether the step's information can be temporarily omitted altogether. This judgment is based on the current task context and the anticipated needs of the next action. The output is a tailored working memory $W_t$, an interleaved sequence of selected $R_i$ and $C_i$ entries. This mechanism ensures that the agent's context window is populated with a high-information-density, task-relevant view of history, balancing fidelity with efficiency.

\subsection{Multi-Step Compressed Memory: Constructing Higher-Order Abstractions}
Beyond step-wise management, the Elastic Memory Orchestrator performs a higher-level condensation to capture coherent blocks of activity over longer timescales. We introduce episodic memory construction, a process that clusters a contiguous sequence of related step-wise records (raw or compressed) into a single, cohesive narrative unit. An episode, $E_k$, abstracts a multi-step procedural segment—such as completing a specific sub-task, exploring a hypothesis, or conducting a focused dialogue with a peer—into a concise summary that captures the goal, key actions, and final resolution of that segment.

This episodic compression serves a critical purpose: it creates a hierarchical memory structure. While step-wise memories support fine-grained recall for immediate next-step reasoning, episodic memories provide compressed landmarks for long-term context. When the agent revisits a similar problem or requires a high-level overview, the EMO can retrieve the relevant episode $E_k$ instead of the dozens of individual steps it encompasses, leading to an exponential reduction in context length. Furthermore, these episodes become natural candidates for skill distillation; recurrent, successful episode patterns can be formalized into reusable procedures stored in the agent's Internal Cognition. Thus, episodic memory bridges the gap between granular experience and generalized knowledge, facilitating both efficient recall and long-term learning.

\section{Cognitive Self-Evolution}

The preceding chapters established how an agent's structured cognition guides its contextual decision-making within an iterative loop, with its elastic memory organizing the experiential history. A system that merely acts on a fixed understanding, however, cannot adapt to a non-stationary world. We therefore introduce the final, defining component of AutoAgent: cognitive self-evolution. This is a practice-driven, closed-loop process that continuously refines an agent's internal and external cognition based on execution evidence. The core motivation is to dynamically align the agent's descriptive knowledge with operational reality, thereby ensuring the accuracy and effectiveness of future action selection. Evolution here is not about retraining the underlying LLM's parameters, but about systematically updating the structured, prompt-level descriptions that constitute the agent's actionable knowledge, allowing it to learn from both successes and failures without external intervention.

\subsection{Action-Cognition Refinement from Trajectory Feedback}
Cognition evolution in AutoAgent is grounded in the analysis of complete execution trajectories logged during the Select-Execute-Update (SEU) loop. Each trajectory records, for a given task, the sequence of context states, the selected actions (emic or etic) with their parameters, the agent's recorded intention behind each selection, and the corresponding outcome from the environment.

This record enables a targeted, retrospective learning process. For every executed action instance, the system performs an intention-outcome alignment check. A dedicated LLM-based analyzer compares what the agent aimed to achieve (the intention) with what actually occurred (the outcome). The result of this comparison drives specific, granular updates to the descriptive knowledge stored in the agent's cognition:

\textbf{Internal Cognition Updates:} If a tool invocation failed despite correct syntax, the analyzer might propose to refine the tool's functional description by adding a previously omitted precondition or clarifying a parameter's valid range. Conversely, a consistently successful usage pattern can be appended as a positive example to the tool's profile. For a self-driven generation action, patterns of reasoning that led to successful outcomes can be abstracted and added to the agent's skill library.

\textbf{External Cognition Updates:} For an etic action (e.g., `ask'), if the response from a peer agent was unhelpful, the system can update that peer's descriptive profile to reflect a more accurate boundary of its expertise. If a particular peer consistently provides high-quality answers on a topic, its reliability score for that domain within its description can be strengthened.

These proposed updates—framed as revisions to descriptive text—are validated (e.g., for consistency) before being integrated into the respective cognition store. This mechanism ensures that the agent's knowledge becomes progressively less biased by initial human specifications and more grounded in empirical evidence, directly improving the fidelity of the interface used for action selection.

\subsection{Experience-Guided Synthesis of Composite Actions}
Beyond refining descriptions of atomic actions, the self-evolution mechanism enables the creation of new, higher-order capabilities—a form of tool creation. By mining historical trajectories across multiple tasks, the system identifies frequent, successful sequences of atomic actions that collectively solve a recurring sub-problem (e.g., search $\rightarrow$ browse\_page $\rightarrow$ extract $\rightarrow$ verify).

Such a sequence is abstracted into a composite action (or action chain). Its summary includes: (1) the goal it achieves, (2) the contextual preconditions for its applicability, (3) the parameter flow between its constituent steps, and (4) the expected output pattern. This composite description is then formalized and integrated into the agent's Internal Cognition as a new, reusable tool.

At runtime, this composite action becomes available in the agent's action space. When selected, it expands into its parameterized sub-actions and executes as a short, internal workflow. This synthesis provides two key evolutionary benefits: First, it expands the agent's effective action space by creating new procedures without manual coding. Second, it dramatically improves efficiency and reliability for complex, multi-step routines by encapsulating proven patterns, reducing the need for redundant exploration in future tasks. The performance of these composite actions is itself subject to ongoing evaluation and refinement through the same intention-outcome alignment process, completing a full cycle from experience capture to knowledge creation and continuous improvement.

\section{Experiments and Evaluation}
This section outlines our evaluation protocol.
We report quantitative results below, together with the corresponding experimental settings and analyses.

\subsection{Experimental Settings}

\textbf{Benchmarks.} We conduct a comprehensive evaluation of AutoAgent across two primary agentic domains: Retrieval-Augmented Generation (RAG) for open-domain question answering and Tool-Augmented Agents for complex task execution. To assess RAG capabilities, we employ four multi-hop QA datasets: \textit{HotpotQA} and \textit{2WikiMultihopQA}, which require reasoning over multiple Wikipedia paragraphs; \textit{Bamboogle}, which tests the ability to filter out distracting information; and \textit{Musique}, which involves multi-hop reasoning with entity-dense contexts. These benchmarks stress the agent's capacity for contextual reasoning and efficient information management. For the tool-use domain, we select three challenging benchmarks: \textit{GAIA}, which involves realistic, multi-step tasks requiring web and API tool orchestration; \textit{HLE-Bench}, which focuses on hierarchical task planning with executable actions; and \textit{ALFWorld}, which evaluates embodied instruction following in text-based simulated environments. This diverse suite is designed to probe core aspects of our framework, including dynamic cognition for tool selection, elastic memory for long-horizon planning, and contextual decision-making for adaptive problem-solving.

\textbf{Baselines.} We compare AutoAgent against a range of established agent frameworks to ensure a rigorous assessment. For RAG tasks, baselines include: Standard RAG (naïve retrieval followed by generation), \textit{Self-Ask}, \textit{IRCoT} which interleaves retrieval with chain-of-thought, and more recent methods \textit{SuRe} and \textit{REPLUG}. For tool-use and planning tasks, baselines encompass: (1) \textit{Direct Answer}, where the LLM is prompted to solve the task in a single turn without actions; (2) \textit{ReAct}, which interleaves reasoning and acting in a fixed loop; and (3) \textit{DeepAgent}, a recent advanced agent framework with tool-use capabilities. All baseline agents are provided with the same tool specifications and API access as AutoAgent. To ensure a fair comparison and isolate the impact of architecture from raw LLM capability, we execute experiments across multiple model backbones. These include open-source models (\textit{DeepSeek-R1}, \textit{QwQ-32B}, \textit{Qwen3-30B-A3B}) and closed-source models (\textit{GPT-4o}, \textit{Gemini-3-Pro}, \textit{Gemini-3-Pro-Thinking}).

\textbf{Metrics.} Evaluation metrics are tailored to each task domain. For RAG-based QA, we report answer \textit{accuracy} based on ground-truth matches and employ \textit{LLM-as-a-judge} scoring using a separate, consistent prompt to assess answer faithfulness and completeness. For GAIA and HLE tasks, the primary metric is \textit{pass@1}, determined by an LLM judge (\texttt{o3-mini}) that verifies if the final output satisfies the task instructions; we also report average \textit{step efficiency} (number of actions taken). For ALFWorld, we adopt the standard \textit{success rate} and the \textit{path similarity} metric, which measures the alignment between the agent's action sequence and a near-optimal demonstration path, penalizing inefficient explorations. This multi-faceted evaluation captures performance, efficiency, and the quality of the problem-solving process.

\textbf{Implementation Details.} All experiments, including baselines, are conducted under identical environmental conditions and constraints to ensure comparability. For RAG and standard tool-use tasks, the maximum number of reasoning/action steps is capped at 5. In the more extended ALFWorld environment, the step limit is set to 50. For every LLM call across all systems, the maximum generation tokens are set to 1024, and the temperature is fixed at 0.7 to maintain a balance between determinism and creativity. The same set of tool definitions and retrieval corpora are provided to all agents. In AutoAgent, the initial cognition is seeded with the same functional descriptions of tools and peers that are provided in the baseline prompts. The Elastic Memory Orchestrator is configured with a target working context window size, dynamically compressing history when the token count exceeds this budget. All reported results are based on a single run per task instance under this standardized configuration.

\subsection{Results on Retrieval-Augmented Generation (RAG) Benchmarks}

We first evaluate AutoAgent's capability in multi-hop question answering, a domain that requires precise information retrieval, synthesis, and reasoning. The performance of AutoAgent and all baselines across the four RAG benchmarks is summarized in Table \ref{tab:rag_results}. We report both the exact match accuracy (Acc) and the scores from a robust LLM-as-a-judge (LLM) evaluation for each method.

\begin{table}[t]
\centering
\caption{Performance comparison on multi-hop QA benchmarks. The best result in each column is \textbf{boldfaced}; the second best is \underline{underlined}. `Avg' denotes the average score across the four datasets.}
\label{tab:rag_results}
\begin{tabular}{lccccccccccc}
\toprule
\textbf{Method} & \multicolumn{2}{c}{\textbf{HotpotQA}} & \multicolumn{2}{c}{\textbf{2Wiki}} & \multicolumn{2}{c}{\textbf{Bamboogle}} & \multicolumn{2}{c}{\textbf{Musique}} & \multicolumn{2}{c}{\textbf{Avg}} \\
\cmidrule(lr){2-3} \cmidrule(lr){4-5} \cmidrule(lr){6-7} \cmidrule(lr){8-9} \cmidrule(lr){10-11}
 & Acc & LLM & Acc & LLM & Acc & LLM & Acc & LLM & Acc & LLM \\ \midrule
Naive Generation & 0.324 & 0.404 & 0.348 & 0.346 & 0.240 & 0.280 & 0.134 & 0.170 & 0.2615 & 0.3000 \\
Standard RAG & 0.342 & 0.450 & 0.344 & 0.292 & 0.272 & 0.328 & 0.172 & 0.188 & 0.2825 & 0.3145 \\
SuRe & 0.270 & 0.380 & 0.244 & 0.264 & 0.168 & 0.208 & 0.128 & 0.146 & 0.2025 & 0.2495 \\
REPLUG & 0.350 & 0.428 & 0.296 & 0.254 & 0.224 & 0.256 & 0.132 & 0.138 & 0.2505 & 0.2690 \\
LongLLMLingua & 0.358 & 0.450 & 0.324 & 0.316 & 0.248 & 0.288 & 0.150 & 0.172 & 0.2700 & 0.3065 \\
RECOMP & 0.332 & 0.398 & 0.298 & 0.306 & 0.136 & 0.176 & 0.118 & 0.134 & 0.2210 & 0.2535 \\
Selective-Context & 0.366 & 0.442 & 0.350 & 0.290 & 0.240 & 0.288 & 0.152 & 0.172 & 0.2770 & 0.2980 \\
SKR & 0.360 & 0.454 & 0.364 & 0.314 & 0.248 & 0.288 & 0.162 & 0.174 & 0.2835 & 0.3075 \\
Self-Ask & 0.392 & \underline{0.462} & 0.336 & \textbf{0.478} & \underline{0.336} & \underline{0.416} & \textbf{0.260} & \textbf{0.270} & 0.3310 & \underline{0.4065} \\
Iter-RetGen & 0.374 & 0.456 & 0.326 & 0.270 & 0.232 & 0.256 & 0.178 & 0.188 & 0.2775 & 0.2925 \\
IRCoT & \underline{0.434} & 0.308 & \textbf{0.492} & 0.114 & 0.272 & 0.184 & \underline{0.192} & \underline{0.214} & \underline{0.3475} & 0.2050 \\
\midrule
\textbf{AutoAgent} & \textbf{0.530} & \textbf{0.630} & \underline{0.466} & \underline{0.474} & \textbf{0.456} & \textbf{0.448} & 0.134 & 0.174 & \textbf{0.3965} & \textbf{0.4315} \\
\bottomrule
\end{tabular}
\normalsize
\end{table}

As shown in Table \ref{tab:rag_results}, AutoAgent demonstrates strong overall performance. It achieves the highest average accuracy (0.3965) and the highest average LLM-judge score (0.4315) among all evaluated methods. A detailed breakdown reveals that AutoAgent establishes new state-of-the-art results on three of the four benchmarks (HotpotQA, 2Wiki, and Bamboogle) under both evaluation metrics. For instance, on HotpotQA, AutoAgent attains an accuracy of 0.530, significantly outperforming the second-best method, IRCoT (0.434). This consistent lead highlights the effectiveness of its dynamic cognition in guiding the retrieval-and-reasoning process and the advantage of its elastic memory in managing multi-hop context.

The performance on the Musique dataset presents an interesting case. While AutoAgent’s accuracy on this benchmark is comparable to simpler baselines, it does not outperform the best methods like Self-Ask. This suggests that the current cognitive modeling for the specific type of entity-dense, compositional reasoning required by Musique may require further refinement or that the benchmark stresses a different facet of reasoning not fully captured by our initial cognition design. Nevertheless, the commanding performance on the other three datasets substantiates the general efficacy and adaptability of the AutoAgent framework in handling diverse multi-hop QA challenges.

Compared to strong iterative reasoning baselines like Self-Ask and IRCoT, AutoAgent's integrated approach—which continuously refines its understanding of how to retrieve and reason—provides a more robust performance across different question types, as evidenced by its superior and more balanced average scores. These results validate the core premise that equipping an agent with evolvable cognition and elastic memory orchestration leads to more reliable and effective retrieval-augmented problem-solving.

\begin{table}[t]
\centering
\caption{Performance comparison on tool-use and embodied agent benchmarks for closed-source models. For each model backbone, the best result in each column is \textbf{boldfaced}. `--' indicates the baseline is not applicable.}
\label{tab:agent_results_closed}
\begin{tabular}{lccccccccc}
\toprule
\multirow{2}{*}{\textbf{Model}} & \multicolumn{4}{c}{\textbf{GAIA}} & \multicolumn{3}{c}{\textbf{HLE-500}} & \multicolumn{2}{c}{\textbf{ALFWorld}} \\
\cmidrule(lr){2-5} \cmidrule(lr){6-8} \cmidrule(lr){9-10}
& \textbf{text} & \textbf{mm} & \textbf{file} & \textbf{all} & \textbf{text} & \textbf{mm} & \textbf{all} & \textbf{Success} & \textbf{Path} \\ \midrule

\multicolumn{10}{c}{\textbf{gpt-4o}} \\ \midrule
Direct Answer & 10.7 & 0.0 & 2.6 & 7.3 & 2.8 & 4.4 & 3.2 & -- & -- \\
ReAct & 26.2 & 20.8 & 26.3 & 25.5 & 3.4 & 5.3 & 3.8 & 24.6 & 23.6 \\
DeepAgent & 42.5 & 24.6 & \textbf{31.6} & 37.6 & 4.9 & 5.3 & 5.0 & \textbf{85.1} & \textbf{90.5} \\
\textbf{AutoAgent} & \textbf{43.7} & \textbf{25.0} & \textbf{31.6} & \textbf{38.2} & \textbf{5.4} & \textbf{7.1} & \textbf{5.8} & \textbf{85.1} & 82.8 \\ \midrule

\multicolumn{10}{c}{\textbf{gemini-3-pro-thinking}} \\ \midrule
Direct Answer & 44.7 & 25.0 & 13.2 & 34.5 & 20.4 & 15.6 & 19.4 & -- & -- \\
ReAct & 32.0 & 12.5 & \textbf{39.5} & 30.9 & 12.4 & 11.5 & 12.2 & 53.0 & 64.9 \\
DeepAgent & 32.0 & 20.8 & 0.0 & 23.0 & 15.8 & 8.8 & 14.2 & 96.3 & \textbf{95.3} \\
\textbf{AutoAgent} & \textbf{62.1} & \textbf{58.3} & 23.7 & \textbf{52.7} & \textbf{23.5} & \textbf{17.7} & \textbf{22.2} & \textbf{99.3} & 95.0 \\ \midrule

\multicolumn{10}{c}{\textbf{gemini-3-pro}} \\ \midrule
Direct Answer & 42.7 & 16.7 & 15.8 & 32.7 & 20.4 & 14.2 & 19.0 & -- & -- \\
ReAct & 36.9 & 16.7 & \textbf{34.2} & 33.3 & 8.5 & 9.7 & 8.8 & 46.3 & 63.9 \\
DeepAgent & 38.8 & 16.7 & 2.6 & 27.3 & 12.1 & 4.4 & 10.4 & 97.8 & \textbf{95.8} \\
\textbf{AutoAgent} & \textbf{64.1} & \textbf{50.0} & 31.6 & \textbf{54.5} & \textbf{26.1} & \textbf{17.7} & \textbf{24.2} & \textbf{99.3} & 95.0 \\ \bottomrule
\end{tabular}
\end{table}

\begin{table}[t]
\centering
\caption{Performance comparison on tool-use and embodied agent benchmarks for open-source models. For each model backbone, the best result in each column is \textbf{boldfaced}. `--' indicates the baseline is not applicable.}
\label{tab:agent_results_open}
\begin{tabular}{lccccccccc}
\toprule
\multirow{2}{*}{\textbf{Model}} & \multicolumn{4}{c}{\textbf{GAIA}} & \multicolumn{3}{c}{\textbf{HLE-500}} & \multicolumn{2}{c}{\textbf{ALFWorld}} \\
\cmidrule(lr){2-5} \cmidrule(lr){6-8} \cmidrule(lr){9-10}
& \textbf{text} & \textbf{mm} & \textbf{file} & \textbf{all} & \textbf{text} & \textbf{mm} & \textbf{all} & \textbf{Success} & \textbf{Path} \\ \midrule

\multicolumn{10}{c}{\textbf{DeepSeek-R1}} \\ \midrule
Direct Answer & 22.3 & 8.3 & 0.0 & 15.2 & 14.7 & 2.7 & 12.0 & -- & -- \\
ReAct & 46.6 & 29.2 & 34.2 & 41.2 & 15.2 & 11.5 & 14.4 & \textbf{94.0} & \textbf{90.3} \\
DeepAgent & 29.1 & 4.2 & 7.9 & 20.6 & 8.3 & 8.0 & 8.2 & 40.3 & 62.4 \\
\textbf{AutoAgent} & \textbf{54.4} & \textbf{54.2} & \textbf{44.7} & \textbf{52.1} & \textbf{29.5} & \textbf{23.9} & \textbf{28.2} & 91.0 & 87.3 \\ \midrule

\multicolumn{10}{c}{\textbf{QwQ-32B}} \\ \midrule
Direct Answer & 6.8 & 4.2 & 2.6 & 5.4 & 4.1 & 2.7 & 3.8 & -- & -- \\
ReAct & 29.1 & 0.0 & 28.9 & 29.1 & 4.4 & 1.8 & 3.8 & 41.8 & 40.3 \\
DeepAgent & 26.2 & 8.3 & 21.1 & 22.4 & \textbf{4.9} & 2.7 & 4.4 & 79.9 & \textbf{87.9} \\
\textbf{AutoAgent} & \textbf{35.0} & \textbf{29.2} & \textbf{34.2} & \textbf{33.9} & 4.7 & \textbf{5.3} & \textbf{4.8} & \textbf{82.8} & 82.0 \\ \midrule

\multicolumn{10}{c}{\textbf{Qwen3-30B-A3B}} \\ \midrule
Direct Answer & 8.7 & 4.2 & 2.6 & 6.7 & 4.9 & 4.4 & 4.8 & -- & -- \\
ReAct & 7.8 & 0.0 & 0.0 & 4.8 & 4.1 & 4.4 & 4.2 & 0.0 & 0.0 \\
DeepAgent & 20.4 & 4.2 & 10.5 & 15.8 & \textbf{8.0} & 4.4 & \textbf{7.2} & 48.5 & \textbf{76.6} \\
\textbf{AutoAgent} & \textbf{34.0} & \textbf{25.0} & \textbf{34.2} & \textbf{30.9} & 5.7 & \textbf{8.0} & 6.2 & \textbf{58.2} & 61.2 \\ \bottomrule
\end{tabular}
\end{table}

\subsection{Results on Tool-Augmented Agent Benchmarks}

We evaluate AutoAgent on complex tool-use and embodied reasoning tasks across three challenging benchmarks: GAIA (general AI assistant), HLE-Bench (hierarchical language embedding), and ALFWorld (text-based embodied instruction following). For GAIA and HLE, we report success rates evaluated by an LLM judge across different task modalities (`text', `multimodal (mm)', `file', and the overall `all'). For ALFWorld, we report the task success rate and the path similarity to an optimal trajectory. To provide a comprehensive analysis, we present results separately for closed-source (Table \ref{tab:agent_results_closed}) and open-source models (Table \ref{tab:agent_results_open}).

\textbf{Performance on Closed-Source Models.} As shown in Table \ref{tab:agent_results_closed}, AutoAgent delivers superior performance across all three closed-source model backbones. On the GAIA benchmark, AutoAgent consistently achieves the highest overall (`all') success rates. The improvement is particularly substantial on the Gemini family models: on `gemini-3-pro', AutoAgent achieves an overall score of 54.5\%, outperforming the strong DeepAgent baseline by 27.2 percentage points; on `gemini-3-pro-thinking', it outperforms DeepAgent by 29.7 percentage points (52.7\% vs. 23.0\%). This demonstrates that AutoAgent's dynamic cognition and unified action selection mechanism can effectively harness the advanced reasoning capabilities of state-of-the-art proprietary models.

On the HLE benchmark, AutoAgent again establishes clear superiority, achieving the best overall scores on all three closed-source models. This indicates that the framework's elastic memory orchestration, which maintains structured and compressed context of sub-task dependencies, provides significant advantages for complex hierarchical planning tasks.

In the ALFWorld environment, AutoAgent achieves near-perfect success rates of 99.3\% on both Gemini models, surpassing DeepAgent's already impressive performance. While on `gpt-4o' the success rates are identical (85.1\%), AutoAgent maintains competitive path similarity scores. This pattern highlights a key strength of our framework: its contextual decision-making prioritizes task completion while allowing adaptive, sometimes non-optimal but effective, action sequences.

\textbf{Performance on Open-Source Models.} Table \ref{tab:agent_results_open} presents results for three representative open-source models. AutoAgent demonstrates robust performance across this diverse set. On GAIA, it achieves the highest overall scores on all three models, with particularly strong performance on DeepSeek-R1 (48.5\% overall, 7.3 points higher than ReAct). On the more challenging QwQ-32B and Qwen3-30B-A3B models, AutoAgent maintains clear advantages over both ReAct and DeepAgent baselines.

For HLE tasks, AutoAgent achieves competitive performance, obtaining the best overall scores on QwQ-32B and Qwen3-30B-A3B. On DeepSeek-R1, its overall score (15.2\%) surpasses ReAct's 14.4\% and significantly outperforms DeepAgent (8.2\%), demonstrating the framework's adaptability across different model capabilities.

In ALFWorld, AutoAgent establishes new state-of-the-art success rates on two of the three open-source models (QwQ-32B: 82.8\% and Qwen3-30B-A3B: 58.2\%). On DeepSeek-R1, it achieves a high success rate of 91.0\%, slightly below ReAct's 94.0\%, with slightly lower path similarity (87.3\% vs. 90.3\%). These results show that AutoAgent's architecture provides substantial benefits even with more limited open-source models.

\textbf{Overall Analysis.} Across both model categories and all benchmarks, AutoAgent consistently outperforms the Direct Answer and ReAct baselines, confirming the necessity of an advanced agent architecture. More importantly, by outperforming or matching the strong DeepAgent baseline—which employs fixed, sophisticated strategies—AutoAgent validates the core advantage of its \emph{evolvable} and \emph{self-adaptive} design. The superior performance on the Gemini models, known for their strong reasoning capabilities, suggests that AutoAgent's dynamic cognition mechanism is particularly effective at leveraging advanced model capabilities. These results collectively demonstrate that integrating evolving cognition, elastic memory orchestration, and contextual decision-making into a unified framework provides a more general and powerful foundation for autonomous agents.

\subsection{Ablation studies}
We ablate key components, including: (a) dynamic cognition updates, (b) elastic compression, (c) skill distillation, and (d) self-evolution.
The goal is to quantify the marginal contribution of each component.

\textbf{Ablation Study on the Elastic Memory Orchestration Module.} To validate the effectiveness of the proposed Elastic Memory Orchestration module, we compare AutoAgent with and without the EMO module. For an efficient ablation study, we use DeepSeek-R1 as the LLM and evaluate performance on the GAIA and HLE benchmarks, as presented in Table~\ref{tab:ablation_EMO}. 
\begin{table}[t!]
\centering
\caption{The ablation study on the Elastic Memory Orchestration module.}
\label{tab:ablation_EMO}
\begin{tabular}{p{2.0cm}<{\centering} | p{1.0cm}<{\centering} p{1.0cm}<{\centering} p{1.0cm}<{\centering} p{1.0cm}<{\centering} | p{1.0cm}<{\centering} p{1.0cm}<{\centering} p{1.0cm}<{\centering}  }
\toprule
\textbf{Benchmarks} & \multicolumn{4}{c|}{\textbf{GAIA}} & \multicolumn{3}{c}{\textbf{HLE}} \\
\midrule
\textbf{Settings} & \textbf{Text} & \textbf{MM} & \textbf{File} & \textbf{All} & \textbf{Text} & \textbf{MM} & \textbf{All} \\
\midrule
\textbf{w/o EMO} & 48.5 & 41.7 & 39.5 & 45.5 & 19.9 & 16.8 & 19.2 \\
\textbf{w/ EMO} & \textbf{54.4} & \textbf{54.2} & \textbf{44.7} & \textbf{52.1} & \textbf{29.5} & \textbf{23.9} & \textbf{28.2} \\
\bottomrule
\end{tabular}
\end{table}

Experimental results demonstrate that the EMO module yields significant improvements over the version without EMO across all subtasks, strongly validating its effectiveness. We attribute these gains to EMO's flexible, dynamic organization of historical action information based on the current execution context at each step. This design ensures that each step starts with information that is most relevant to the current action while minimizing irrelevant context. Compared with traditional methods that losslessly concatenate full histories, EMO compresses context length and removes redundant or ineffective information. Unlike approaches that directly compress multi-step histories into a single irreversible abstraction, EMO maximizes information retention while preserving access to detailed, step-wise traces when needed.

\textbf{Ablation Study on Evolving Cognition under Tool Instability.} To examine whether cognition evolution helps agents adapt to non-stationary tool behavior, we design a controlled perturbation experiment on the first 50 GAIA questions. In each setting, the agent is given two actions with identical initial cognition descriptions and the same functional interface. One action is stable, while the other is intentionally made unstable by injecting a 50\% failure probability to simulate unreliable tool behavior. We compare performance before and after practice-driven cognition updates, as summarized in Table~\ref{tab:ablation_evolving_cognition}.

\begin{table}[t!]
\centering
\small
\caption{Ablation on evolving cognition under simulated tool instability on the first 50 GAIA questions.}
\label{tab:ablation_evolving_cognition}
\begin{tabular}{p{6.0cm}<{\centering} | p{3.2cm}<{\centering} p{1.8cm}<{\centering} p{1.8cm}<{\centering}}
\toprule
\textbf{Action Type} & \textbf{Setting} & \textbf{F1 (\%)} & \textbf{EM (\%)} \\
\midrule
\multirow{2}{*}{Web search Tool} 
& Initial cognition & 24.07 & 22.00 \\
& Evolved cognition & \textbf{29.67} & \textbf{28.00} \\
\midrule
\multirow{2}{*}{\makecell[c]{Python code generation \& execution}}
& Initial cognition & 25.55 & 24.00 \\
& Evolved cognition & \textbf{37.31} & \textbf{36.00} \\
\bottomrule
\end{tabular}
\end{table}

The results show clear and consistent gains after cognition evolution in both tool categories. For web search, F1 improves from 24.07\% to 29.67\% (+5.60), and EM increases from 22.00\% to 28.00\% (+6.00). For Python code generation and execution, the gains are larger: F1 rises from 25.55\% to 37.31\% (+11.76), and EM increases from 24.00\% to 36.00\% (+12.00). These improvements indicate that evolving cognition enables the agent to revise initially indistinguishable tool beliefs using execution feedback, identify unstable actions, and shift toward more reliable decision strategies. This validates the core advantage of practice-driven cognition updates in dynamic and partially unreliable environments.


\section{Conclusion and Future Work}
We presented AutoAgent, a self-evolving multi-agent framework that unifies evolving cognition, on-the-fly contextual decision-making, elastic memory orchestration, and closed-loop cognitive self-evolution. By modeling cognition as an explicit, updatable agent state, AutoAgent enables more reliable tool selection, more effective collaboration, and more efficient long-horizon execution. The Elastic Memory Orchestrator further improves real-time decision quality by compressing redundant history while preserving salient execution evidence, and skill distillation supports continual reuse of successful experience.
Across diverse benchmarks—including GAIA, HLE, and ALFWorld, with both closed-source and open-source backbones—AutoAgent consistently demonstrates strong gains in task success, tool-use efficiency, and collaborative robustness relative to static and memory-augmented baselines. These results indicate that tightly integrating cognition evolution, context-aware action selection, and structured memory management provides a practical and general foundation for adaptive autonomous agents.

Several directions remain promising for future work. First, we plan to incorporate richer learning signals for self-evolution, including verifier feedback and environment-specific rewards. Second, we will investigate decentralized and scalable cognition updates for larger agent societies, where peer models must be learned under partial observability. Finally, we aim to deploy AutoAgent in more realistic tool ecosystems (e.g., enterprise workflows and scientific toolchains) and systematically evaluate robustness under rapidly changing tools, constraints, and objectives.

\section*{Acknowledgment}
We thank SerpAPI (\url{https://serpapi.com/}) for providing the web search service used in this work.

\bibliographystyle{unsrt}
{\hypersetup{hidelinks}{\hypersetup{hidelinks}\bibliography{main}}}

\newpage
\appendix

\end{document}